\documentclass{article}

\usepackage{graphicx}        
     
\usepackage{amsmath}
\usepackage{amssymb}

\newcommand{\proscal}[2]{\left\langle#1,#2\right\rangle}
\newcommand{\norme}[1]{\left\|#1\right\|}

\newcommand{\esp}[1]{\mathbb{E}\left[#1\right]}

\newcommand{\V}[1]{\mathrm{Var}\left[#1\right]}

\newcommand{\egdef}{\stackrel{\mbox{\tiny def}}{=}}

\newcommand{\R}{\mathbb{R}}
\newcommand{\Rp}{{\mathbb{R}^p}}
\newcommand{\1}{\mathrm{Id}}
\newtheorem{algorithm}{Algorithm}

\newtheorem{theorem}{Theorem}
\newtheorem{proposition}{Proposition}
\newtheorem{corollary}{Corollary}

\begin{document}

\title{Auto-associative models, nonlinear Principal component analysis, manifolds and projection pursuit}
\author{St\'ephane Girard$^{(1)}$ and Serge Iovleff$^{(2)}$}

\date{INRIA Rh\^one-Alpes, projet Mistis,  Inovall\'ee, 655, av. de l'Europe, Montbonnot, 38334 Saint-Ismier cedex, France.\\
\texttt{Stephane.Girard@inrialpes.fr}\\
Laboratoire Paul Painlev\'e, 
 59655 Villeneuve d'Ascq Cedex, France.\\ \texttt{serge.iovleff@univ-lille1.fr}}

\maketitle

\section{Introduction}

Principal component analysis (PCA) is a well-known method
for extracting linear structures from high-dimensional datasets.
It computes the subspace best approaching the dataset from
the Euclidean point of view.
This method benefits from efficient implementations based either on 
solving an eigenvalue problem or on iterative algorithms.
We refer to~\cite{Jolliffe} for details.
In a similar fashion, multi-dimensional scaling~\cite{Carroll,Kruskal,Shepard}
addresses the problem of finding the linear subspace
best preserving the pairwise distances.
More recently, new algorithms have been proposed to compute
low dimensional embeddings of high dimensional data. 
For instance, Isomap~\cite{ISOMAP}, 
LLE (Locally linear embedding)~\cite{LLE} and CDA
(Curvilinear distance analysis)~\cite{Demartines}
aim at reproducing in the projection
space the structure of the initial local neighborhood. 
These methods are mainly dedicated to visualization purposes.  
They cannot produce an analytic form of the 
transformation function, making it difficult to map new points into
the dimensionality-reduced space. 
Besides, since they rely on local properties of pairwise distances,
these methods are sensitive to noise and outliers.
We refer to~\cite{Lee} for a comparison
between Isomap and CDA and to~\cite{Vlachos} for a comparison
between some features of LLE and Isomap.

Finding nonlinear structures is a challenging problem. 
An important family of methods focuses on self-consistent structures.
The self-consistency concept is precisely defined in~\cite{Flury}.
Geometrically speaking, it means that each point of the structure
is the mean of all points that project orthogonally onto it.
For instance, it can be shown that the $k$-means algorithm~\cite{Hartigan}
 converges to a set of $k$ self-consistent points.
Principal curves and surfaces~\cite{DEL01,HAS89,LEB94,TIB92}
are examples of one-dimensional and two-dimensional self-consistent
structures.  Their practical computation 
requires to solve a nonlinear optimization problem.
The solution is usually non robust and suffers from a high
estimation bias. In~\cite{Kegl}, a polygonal algorithm is proposed
to reduce this bias.
Higher dimensional self-consistent structures are often 
referred to as self-consistent manifolds even though their existence
is not guaranteed for arbitrary datasets. 
An estimation algorithm based on a grid approximation is proposed in~\cite{Gorban}. The fitting criterion involves two smoothness penalty terms
describing the elastic properties of the manifold.

In this paper, auto-associative models are
proposed as candidates to the generalization of PCA. 
We show in paragraph~\ref{manif} that these models
are dedicated to the approximation of the dataset by a manifold.
Here, the word "manifold" refers to the topology properties
of the structure~\cite{Milnor}.
The approximating manifold is built by a
projection pursuit algorithm presented in paragraph~\ref{algo}.
At each step of the algorithm, the dimension of the manifold 
is incremented. Some theoretical properties are provided
in paragraph~\ref{theo}. In particular, we can show that,
at each step of the algorithm, the mean residuals norm is
not increased. Moreover, it is also established that the
algorithm converges in a finite number of steps.
Section~\ref{exemples} is devoted to the presentation of some particular
auto-associative models. They are compared to the classical PCA
and some neural networks models.
Implementation aspects are discussed in Section~\ref{impl}. 
We show that, in numerous cases, no optimization procedure is
required.
Some illustrations on simulated and real data are presented in Section~\ref{secillus}.

\section{Auto-associative models}
\label{definition}

In this chapter, for each unit vector $a\in\Rp$,
we denote by $P_a(.)=\proscal{a}{.}$ the linear projection
from $\Rp$ to $\R$.
Besides, for all set $E$, the identity function $E\to E$
is denoted by $\1_E$.

\subsection{Approximation by manifolds}
\label{manif}

A function $F^d$:~$\Rp \to \Rp$ is a $d$-dimensional auto-associative
function if there exist $d$ unit orthogonal vectors 
$a^k$, called principal directions,
and $d$ continuously differentiable functions $s^k$:~$\R\to \Rp$, 
called regression functions,
such that
\begin{equation}
\label{condi}
P_{a^j}\circ s^k = \delta_{j,k}\1_{\R}\;\mbox{ for all }\; 1\leq j\leq k\leq d,
\end{equation}
where $\delta_{j,k}$ is the Kronecker symbol
and
\begin{equation}
\label{eqfd}
F^d   =   \left(\1_\Rp-s^d \circ P_{a^d}\right) \circ
 \ldots \circ \left(\1_\Rp-s^1 \circ P_{a^1}\right)
  =  \coprod_{k=d}^1 \left(\1_\Rp-s^k \circ P_{a^k}\right).
\end{equation}
The main feature of auto-associative functions is mainly a consequence of~(\ref{condi}):
\begin{theorem}
The equation $F^d(x)=0$, $x\in\Rp$
defines a differentiable $d$-dimensional manifold of $\Rp$.
\end{theorem}
We refer to~\cite{CS} for a proof.
Thus, the equation $F^d(x)=0$ defines a space in which every point has a 
neighborhood which resembles the Euclidean space $\R^d$,
but in which the global structure may be more complicated.
As an example, on a $1$-dimensional manifold, every point has a neighborhood
that resembles a line. In a $2$-manifold, every point has a neighborhood that looks like a plane. Examples include the sphere or the surface of a torus.

Now, let $X$ be a square integrable random vector of $\Rp$.
Assume, without loss of generality, that $X$ is centered
and introduce $\sigma^2(X)\egdef \esp{\norme{X}^2}$.
For all auto-associative function $F^d$, let us consider
$\varepsilon=F^d(X)$. Note that, from the results of Subsection~\ref{theo}
below, $\varepsilon$ is necessarily a centered random vector. 
In this context, $\sigma^2(\varepsilon)$ is called the residual variance.
Geometrically speaking, the realizations of the random vector $X$
are approximated by the manifold $F^d(x)=0$, $x\in\Rp$
and $\sigma^2(\varepsilon)$ represents the variance of $X$ "outside" the manifold.

Of course, such random vector $X$ always satisfies a $0$-dimensional 
auto-associative model with $F^0=\1_{\Rp}$ and 
$\sigma^2(\varepsilon)=\sigma^2(X)$.
Similarly, $X$ always satisfies a $p$-dimensional auto-associative model
with $F^p=0$ and $\sigma^2(\varepsilon)=0$.
In practice, it is important to find a balance between
these two extreme cases by constructing a $d$-dimensional model 
with $d\ll p$ and $\sigma^2(\varepsilon)\ll \sigma^2(X)$.
For instance, in the case where 
the covariance matrix $\Sigma$ of $X$ is of rank~$d$, then $X$
is located on a $d$-dimensional linear subspace defined by the equation
$F^d_{PCA}(x)=0$ with
\begin{equation}
\label{expanlin}
F^d_{PCA}(x)=x-\sum_{k=1}^d P_{a^k}(x)a^k,
\end{equation}
and where $a^k$, $k=1,\dots,d$ are the eigenvectors of $\Sigma$
associated to the positive eigenvalues. A little algebra
shows that~(\ref{expanlin}) can be rewritten as $F^d(x)=0$,
where $F^d$ is a $d$-dimensional auto-associative function
with linear regression functions $a^k(t)=t a^k$ for
$k=1,\dots,d$. Moreover, we have $\sigma^2(\varepsilon)=0$.
Since~(\ref{expanlin}) is the model produced
by a PCA, it straightforwardly follows that PCA is a special
(linear) case of auto-associative models.
In the next section, we propose an 
algorithm to build auto-associative models
with non necessarily linear regression functions,
small dimension and small residual variance. 
Such models could also 
be called "semi-linear" or "semi-parametric" 
since they include a linear/parametric part through the use
of linear projection operators and a non-linear/non-parametric
part through the regression functions.

\subsection{A projection pursuit algorithm}
\label{algo}

Let us recall that, given an unit vector $a\in\Rp$, an index
$I$:~$\R\to\R$ is a
functional measuring the interest of the projection $P_a(X)$
with a non negative real number. 
The meaning of the word "interest" depends on the considered data
analysis problem.
For instance, a possible choice of $I$ is the projected variance
$I\circ P_a(.) =\V{P_a(.)}$.
Some other examples are presented in Section~\ref{choixindex}.
Thus, the maximization of $I\circ P_a(X)$ with respect to $a$
yields the most interesting direction for this given criteria.
An algorithm performing such an optimization is called a 
projection pursuit algorithm.
We refer to~\cite{Huber} and~\cite{Jones}
for a review on this topic.

Let $d\in\{0,\dots,p\}$, and consider the following algorithm
which consists in
applying iteratively the following steps: [A]
computation of the Axes, [P] Projection, [R] Regression and [U]
Update: 
\begin{algorithm}
\label{algoAA}
Define $R^0=X$.\\
For $k=1,\dots,d$:
\begin{itemize}
\item[\rm{[A]}] Determine $a^k=\arg
  \displaystyle\max_{x\in\Rp} I\circ P_x (R^{k-1})$
s.t. $\norme{x}=1$, $P_{a^j}(x)=0,\; 1\leq j<k$.
\item[\rm{[P]}] Compute $Y^k = P_{a^k}(R^{k-1})$.
\item[\rm{[R]}] Estimate $s^k(t)=\esp{R^{k-1}|Y^k=t}$,
\item[\rm{[U]}] Compute $R^k=R^{k-1}-s^k(Y^k)$.
\end{itemize}
\end{algorithm}
The random variables $Y^k$ are called principal variables
and the random vectors $R^k$ residuals. 
Step [A] consists in computing an axis
orthogonal to the previous ones and maximizing a given
index $I$.
Step [P] consists in
projecting the residuals on this axis to determine the
principal variables, and step [R] is devoted to the estimation 
of the regression function of the principal variables best approximating the
residuals. 
Step [U] simply consists in updating the residuals.
Thus, Algorithm~\ref{algoAA} can be seen as a projection pursuit
regression algorithm~\cite{Fried81,Klinke}
since it combines a projection pursuit
step [A] and a regression step [R].
The main problem of such approaches is to define an efficient
way to iterate from $k$ to $k+1$. Here,
the key property is that 
the residuals $R^k$ are orthogonal to the axis $a^k$ since
\begin{eqnarray}
\nonumber P_{a^k}(R^k)&=&P_{a^k}(R^{k-1})-P_{a^k}\circ s^k(Y^k)\\
\nonumber &=& P_{a^k}(R^{k-1})-\esp{P_{a_k}(R^{k-1})|Y^k}\\
\nonumber &=&Y^k-\esp{Y^k|Y^k}\\
\label{eqbase}
&=&0.
\end{eqnarray}
Thus, it is natural to iterate the model construction in
the subspace orthogonal to $a^k$, see the orthogonality
constraint in step [A]. The theoretical results
provided in the next paragraph are mainly consequences
of this property. 

\subsection{Theoretical results}
\label{theo}

Basing on~(\ref{eqbase}), it is easily shown by induction
that both the residuals and the regression functions
computed at the iteration $k$ are almost surely (a.s.) orthogonal to 
the axes computed before. More precisely, one has
\begin{eqnarray}
\label{eq1}
\proscal{a^j}{R^k}=0, \; a.s. &\mbox{ for all }& 1\leq j\leq k\leq d,\\
\label{eq2}
\proscal{a^j}{s^k(Y^k)}=0, \; a.s. &\mbox{ for all }& 1\leq j<k\leq d.
\end{eqnarray}
Besides, the residuals, principal variables and regression functions
are centered:
$$
\esp{R^k}=\esp{Y^k}=\esp{s^k(Y^k)}=0,
$$
for all $1\leq k\leq d$. Our main result is the following:
\begin{theorem}
\label{th-general}
Algorithm~\ref{algoAA} builds a $d$-dimensional auto-associative model
with principal directions $\{a^1,\dots,a^d\}$,
regression functions $\{s^1,\dots,s^d\}$ and residual
$\varepsilon =R^d$. Moreover, one has the expansion
\begin{equation}
\label{expanX}
X=\sum_{k=1}^d s^k(Y^k)+ R^{d},
\end{equation}
where the principal variables $Y^k$ and $Y^{k+1}$ are centered and non-correlated
for $k=1,\dots,d-1$.
\end{theorem}
The proof is a direct consequence of the orthogonality
properties~(\ref{eq1}) and~(\ref{eq2}).
Let us highlight that, for $d=p$, expansion~(\ref{expanX}) yields
an exact expansion of the random vector $X$ as:
$$
X=\sum_{k=1}^p s^k(Y^k),
$$
since $R^p=0$ (a.s.) in view of~(\ref{eq1}).
Finally, note that the approximation properties of the conditional
expectation entails that the sequence of the residual norms is 
almost surely non increasing.
As a consequence, the following corollary will prove useful to select
the model dimension similarly to the PCA case.
\begin{corollary}
\label{QD} Let $Q_d$ be the information ratio represented by
the $d$-dimensional auto-associative model:
$$
Q_d=1 - \sigma^2(R^d) \left/ \sigma^2(X) \right..
$$
Then, $Q_0=0$, $Q_p=1$ and the sequence $(Q_d)$ is non decreasing.
\end{corollary}
Note that all these properties are quite general, since
they do not depend either on the index $I$, nor on the
estimation method for the conditional expectation.
In the next section, we show how, in particular cases,
additional properties can be obtained.

\section{Examples}
\label{exemples}

We first focus on the auto-associative models which can be
obtained using linear estimators of the regression functions.
The existing links with PCA are highlighted.
Second, we introduce the intermediate class of additive
auto-associative models and compare it to some neural network
approaches.

\subsection{Linear auto-associative models and PCA}

Here, we limit ourselves to linear estimators of the conditional
expectation in step [R].
\noindent At iteration $k$, we thus assume 
$$
s^k(t)=t b^k, \; t\in\R\;,
b^k\in\Rp.
$$
Standard optimization arguments (see~\cite{JMVA}, 
Proposition~2) shows that, necessarily, the regression
function obtained at step [R] is located on the axis
\begin{equation}
\label{bk}
b^k={\Sigma_{k-1} a^k}/({^t a^k \Sigma_{k-1} a^k}),
\end{equation}
with $\Sigma_{k-1}$ the covariance matrix of $R^{k-1}$:
\begin{equation}
\label{sigma}
\Sigma_{k-1}=\esp{R^{k-1}\,^tR^{k-1}},
\end{equation}
and where, for all matrix $M$, the transposed matrix is denoted by $^t M$.
As a consequence of Theorem~\ref{th-general}, we have the
following linear expansion:
$$
X=\sum_{k=1}^d  \frac{Y^k\Sigma_{k-1} a^k}{^t a^k \Sigma_{k-1} a^k} + R^{d}.
$$
As an interesting additional property of these so-called linear
auto-associative models, we have
$\esp{Y_j Y_k}=0$
for all $1\leq j<k\leq d$. This property is established
in~\cite{JMVA}, Proposition~2.  Therefore, the limitation
to a family of linear functions in step [R] allows to recover an important
property of PCA models: the non-correlation of the principal variables.
It is now shown that Algorithm~\ref{algoAA} can also compute a PCA
model for a well suited choice of the index.
\begin{proposition}
If the index in step [A] is the projected variance,
i.e.  
$$
I\circ P_x(R^{k-1})= \V{P_x(R^{k-1})},
$$
and step [R] is given by~(\ref{bk}) then 
Algorithm~\ref{algoAA} computes the PCA model of $X$.
\end{proposition}
Indeed, the solution $a^k$ of step [A] is the
eigenvector associated to the maximum eigenvalue of
$\Sigma_{k-1}$. From~(\ref{bk}) it follows that $b^k=a^k$.
Replacing in~(\ref{eqfd}), we obtain, for orthogonality reasons, 
$F^d=F^d_{PCA}$.

\subsection{Additive auto-associative models and neural networks}

A $d$-dimensional auto-associative function is called additive if~(\ref{eqfd})
can be rewritten as
\begin{equation} 
\label{eqaddi}
F^d= \1_{\R^d} -\sum_{k=1}^d s^k\circ P_{a^k}.
\end{equation}
In~\cite{CRAS}, the following characterization of additive auto-associative functions is provided.
A $d$-dimensional auto-associative function is additive
if and only if 
$$P_{a^j} \circ s^k=\delta_{j,k} \1_{\R} \mbox{ for all }
(j, k)\in\{1,\dots,d\}^2.
$$
As a consequence, we have:
\begin{theorem}
In the linear subspace spanned by $\{a^1,\dots,a^d\}$, every $d$-dimensional
additive auto-associative model reduces to the PCA model.
\end{theorem}
A similar result can be established for the nonlinear PCA based on 
a neural network and introduced in~\cite{KarJou}.
The proposed model is obtained by introducing a nonlinear function
$g:\R\to\R$, called activation function, in the PCA model~(\ref{expanlin}) to obtain
\begin{equation}
\label{eqKJ}
F^d_{KJ}(x)=x-\sum_{k=1}^d g\circ P_{a^k}(x)a^k.
\end{equation}
Note that~(\ref{eqKJ}) is an additive auto-associative model
as defined in~(\ref{eqaddi})
if and only if $g= \1_{\R}$, {\it i.e.} if and only if
it reduces to the PCA model in the linear subspace spanned by $\{a^1,\dots,a^d\}$.
Moreover, in all cases, we have 
$$
\{F^d_{KJ}(x)=0,\;x\in \R^p\} \subset \{F^d_{PCA}(x)=0,\;x\in \R^p\},
$$
which means that this model is included in the PCA one.
More generally, the auto-associative Perceptron with one hidden
layer~\cite{CT} is based on multidimensional activation functions 
$\sigma^k:\R\to\R^p$:
\begin{equation}
\label{eqAAP}
F^d_{AAP}(x)=x-\sum_{k=1}^d \sigma^k\circ P_{a^k}(x).
\end{equation}
Unfortunately, it can be shown~\cite{DK} that a single hidden layer
is not sufficient. Linear activation functions (leading to a PCA)
already yield the best approximation of the data.
In other words, the nonlinearity introduced in~(\ref{eqAAP}) has no significant
effect on the final approximation of the dataset. 
Besides, determining $a^k$, $k=1,\dots,d$
is a highly nonlinear problem with numerous local minima,
and thus very dependent on the initialization.

\section{Implementation aspects}
\label{impl}

In this section, we focus on the implementation aspects
associated to Algorithm~\ref{algoAA}.
Starting from a $n$-sample $\{X_1,\dots,X_n\}$, two problems
are addressed. In Subsection~\ref{estireg}, we propose
some simple methods to estimate the regression functions $s^k$
appearing in step [R]. In Subsection~\ref{choixindex},
the choice of the index in step [A] is discussed. In particular,
we propose a contiguity index whose maximization is explicit.

\subsection{Estimation of the regression functions}
\label{estireg}

\subsubsection{Linear auto-associative models}
To estimate the regression functions, the simplest
solution is to use a linear approach leading to a
linear auto-associative model. In this case, the 
regression axis is explicit, see~(\ref{bk}), and
it suffices to replace $\Sigma_{k-1}$ defined in~(\ref{sigma})
by its empirical counterpart
\begin{equation}
\label{covmat}
V_{k-1}=\frac{1}{n}\sum_{i=1}^n   R^{k-1}_i \,^tR^{k-1}_i,
\end{equation}
where $R^{k-1}_i$ is the residual associated to $X_i$ at
iteration $k-1$.

\subsubsection{Nonlinear auto-associative models}
Let us now focus on nonlinear estimators of the 
conditional expectation $s^k(t)=\esp{R^{k-1}|Y^k=t}$, $t\in\R$.
Let us highlight that $s^k$ is a univariate function
and thus its estimation does not suffer from the curse
of dimensionality~\cite{Bellman}. 
This important property is a consequence of the
"bottleneck" trick used in~(\ref{eqfd}) and, more generally,
in neural networks approaches. The key point is that,
even though $s^k\circ P_{a^k}$ is a $p$- variate function,
its construction only requires the nonparametric estimation
of a univariate function thanks to the projection operator.

For the sake of simplicity, we propose to work in
the orthogonal basis $B^k$ of $\Rp$ obtained by completing 
$\{a^1,\dots,a^{k}\}$. Let us denote by 
$R^{k-1}_j$ the $j$-th coordinate of $R^{k-1}$ in $B^k$.
In view of~(\ref{eq1}), $R^{k-1}_j=0$ for $j=1,\dots,k-1$.
Besides, from step [P], $R^{k-1}_k=Y^k$. Thus,
the estimation of $s^k(t)$ reduces to 
the estimation of $p-k$ functions 
$$
s^k_j(t)= \esp{R^{k-1}_j|Y^k=t},\;\; j=k+1,\dots,p.
$$
This standard problem~\cite{Hardle,LivreToulouse} can be tackled either by
kernel~\cite{Bosq} or projection~\cite{green} estimates.

\paragraph{Kernel estimates}
Each coordinate
$j\in\{k+1,\dots,p\}$ of the estimator can be written in the basis $B^j$ as:
\begin{equation}
\label{noyau}
\hat{s}^k_j(t) = \sum_{i=1}^n R^{k-1}_{j,i}
                                 K\left(\frac{t-Y^k_{i}}{h}\right)
                   \left/ \sum_{i=1}^n K\left(\frac{t-Y^k_{i}}{h}\right) \right.,
\end{equation}
where $R^{k-1}_{j,i}$ represents the $j$-th coordinate of
the residual associated to the observation $X_i$ at the $(k-1)$-th iteration in
the basis $B^k$, $Y^k_{i}$ is the value of the $k$-th
principal variable for the observation $X_i$ and
$K$ is a Parzen-Rosenblatt kernel, that is to say a bounded real
function, integrating to one and such that $t K(t)\to 0$ as
$|t|\to\infty$. For instance, one may use a
a standard Gaussian density.
The parameter $h$ is a positive number called window in this context.
In fact, $\hat{s}^k_j(t)$ can be seen as a weighted mean of the residuals
$R^{k-1}_{j,i}$ which are close to $t$: 
$$
\hat{s}^k_j(t) = \sum_{i=1}^n R^{k-1}_{j,i} w_i^k(t),
$$
\noindent where the weights are defined by
$$
 w_i^k(t)=  K\left(\frac{t-Y^k_{i}}{h}\right)
 \left/ \sum_{i=1}^n K\left(\frac{t-Y^k_{i}}{h}\right) \right.,
$$
and are summing to one:
$$
 \sum_{i=1}^n w_i^k(t)=1.
$$
The amplitude of the
smoothing is tuned by $h$. In the case of a kernel with bounded support,
for instance if supp$(K)=[-1,1]$, the smoothing is performed on
an interval of length $2h$.
For an automatic choice of the smoothing parameter $h$,
we refer to~\cite{Hastie2}, Chapter~6.

\paragraph{Projection estimates}

Each coordinate
$j\in\{k+1,\dots,p\}$ of the estimator is expanded on a basis of $L$ real
functions $\{b_\ell(t),\;\ell=1,\dots,L\}$ as:
$$
\tilde{s}^k_j(t) = \sum_{\ell=1}^L \tilde\alpha_{j,\ell}^k b_\ell(t).
$$
The coefficients $\tilde\alpha_{j,\ell}^k$ appearing in the linear combination
of basis functions are determined such that $\tilde{s}^k_j(Y^k_{i})
\simeq R^{k-1}_{j,i}$ for $i=1,\dots,n$. More precisely,
$$
\tilde\alpha_{j,.}^k =\arg\min_{\alpha_{j,.}^k} \sum_{i=1}^n \left(\sum_{\ell=1}^L \alpha_{j,\ell}^k b_\ell(Y^k_{i})
-  R^{k-1}_{j,i}\right)^2,
$$
and it is well-known that this least-square problem benefits from
an explicit solution which can be matricially written as
\begin{equation}
\label{proj}
\tilde\alpha_{j,.}^k = (^tB^k B^k)^{-1} \,^tB^k  R^{k-1}_{j,.}
\end{equation}
where $B^k$ is the $n\times L$ matrix with coefficients $B^k_{i,\ell}= b_\ell(Y^k_{i})$,
$i=1,\dots n$, $\ell=1,\dots,L$. Note that this matrix does not
depend on the coordinate $j$. Thus, the matrix inversion in~(\ref{proj})
is performed only once at each iteration $k$. Besides, the size
of this matrix is $L\times L$ and thus does not depend either on the dimension
of the space $p$, nor on the sample size $n$.
As an example, one can use a basis of cubic splines~\cite{EU90}.
In this case, the parameter $L$ is directly linked to $N$ the 
number of knots: $L=N+4$. Remark that, in this case, condition $N+4\leq n$
is required so that the matrix is $^tB^k B^k$ is regular.

\subsection{Computation of principal directions}
\label{choixindex}

\noindent The choice of the index $I$ is the key point of any
projection pursuit problem where it is needed to find 
"interesting" directions. We refer to~\cite{Huber} and~\cite{Jones}
for a review on this topic. Let us recall that the
meaning of the word "interesting" depends on the considered data
analysis problem.
As mentioned in Subsection~\ref{algo}, the most popular index is
the projected variance 
\begin{equation}
\label{indexpca}
I_{PCA}\circ P_x\left(R^{k-1}\right) = \frac{1}{n}{\sum_{i=1}^n P_{x}^2(R^{k-1}_i)}
\end{equation}
used in PCA. Remarking that this index can be rewritten as
$$
I_{PCA}\circ P_x\left(R^{k-1}\right) = \frac{1}{2n^2}\sum_{i=1}^n\sum_{j\neq i}
 P_{x}^2(R^{k-1}_i-R_j^{k-1}),
$$
it appears that the "optimal" axis maximizes the mean
distance between the projected points.
An attractive feature of the index~(\ref{indexpca}) is that its
maximization benefits from an explicit solution in terms
of the eigenvectors of the empirical covariance matrix $V_{k-1}$
defined in~(\ref{covmat}).
Friedman {\it et al}~\cite{Fried74, Fried87},
and more recently Hall~\cite{Hall}, proposed an
index to find clusters or use deviation from the normality measures
to reveal more complex structures of the scatter-plot.
An alternative approach can be found in~\cite{Caussinus} where a
particular metric is introduced in PCA so as to detect clusters.
We can also mention indices dedicated to outliers detection~\cite{Pan}. 
Similar problems occur in the neural networks context where
the focus is on the construction of 
nonlinear mappings to unfold the manifold. It is usually
required that such a mapping preserves that local
topology of the dataset.
In this aim, Demartines and Herault~\cite{Demartines} introduce
an index to detect the directions in which the nonlinear projection
approximatively preserves distances. Such an index can be adapted
to our framework by restricting ourselves to linear projections: 
$$
\begin{array}{ll}
I_{DH}\circ P_x\left(R^{k-1}\right) =
\displaystyle\sum_{i=1}^n\sum_{j\neq i}&
\left( \norme{R_i^{k-1}-R_j^{k-1}}-|P_x|(R_i^{k-1}-R_j^{k-1})
\right)^2 \\
& H\circ|P_{x}|(R_i^{k-1}-R_j^{k-1})
.
\end{array}
$$
The function $H$ is assumed to be positive and non increasing
in order to favor the local topology preservation. 
According the authors, the application of this function to
the outputs $P_x R_i^{k-1}$ instead of the inputs $R_i^{k-1}$
allows to obtain better performances than the Kohonen's self-organizing
maps~\cite{Koh1,Koh2}. 
Similarly, the criterion introduced in~\cite{Sammson} yields
in our case
$$
\begin{array}{ll}
I_{S}\circ P_{x}(R^{k-1}) &=
\displaystyle\sum_{i=1}^n\sum_{j\neq i}
\left( \norme{R_i^{k-1}-R_j^{k-1}}-|P_{x}|(R_i^{k-1}-R_j^{k-1})
\right)^2 \\
& \left / \displaystyle\sum_{i=1}^n\sum_{j\neq i}P_{x}^2(R_i^{k-1}-R_j^{k-1}) \right..
\end{array}
$$
However, in both cases, the resulting functions
are nonlinear
and thus difficult to optimize with respect to~$x$.

\begin{figure}
\begin{center}
\includegraphics[height=4cm]{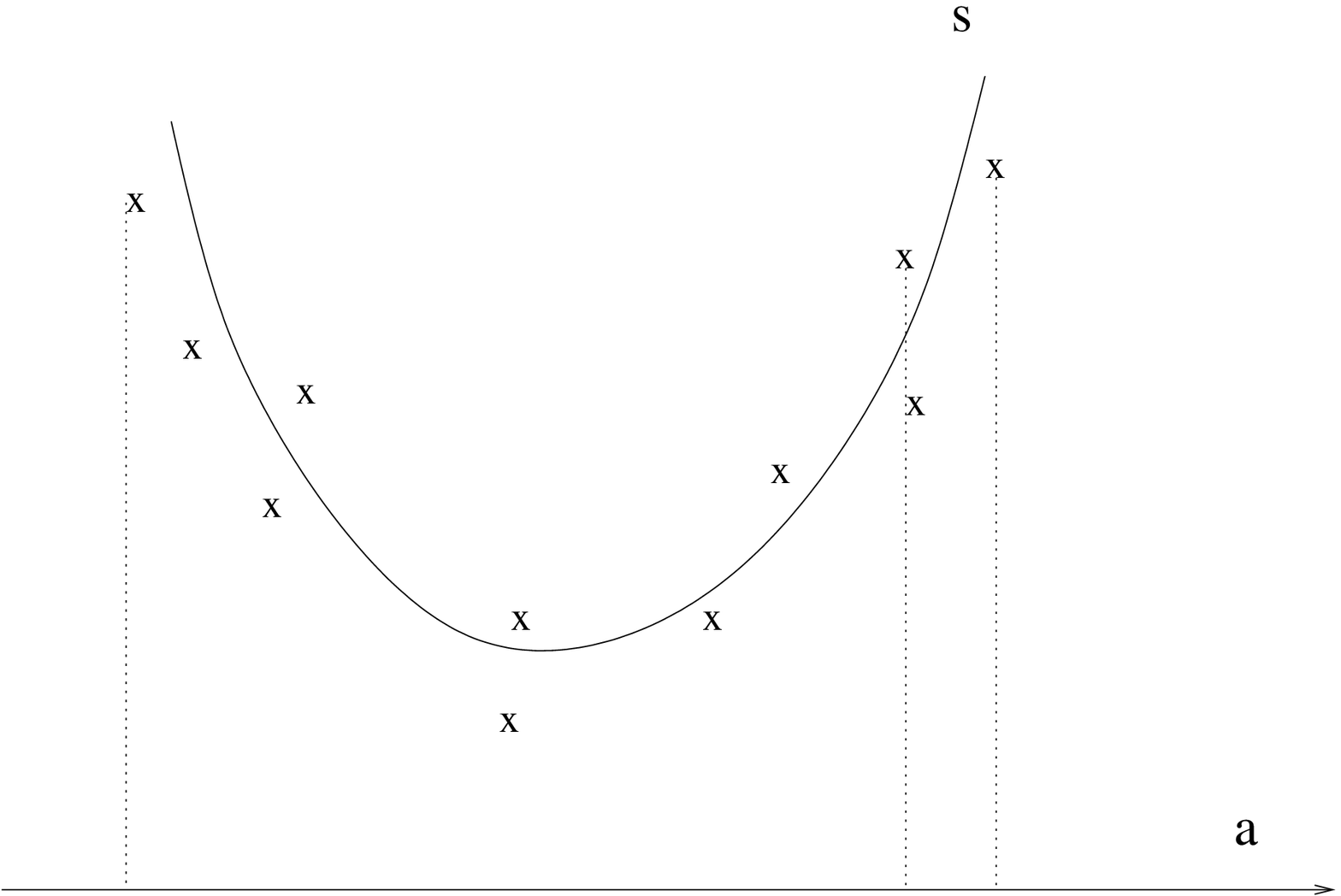}
\hspace*{1cm}
\includegraphics[height=4cm]{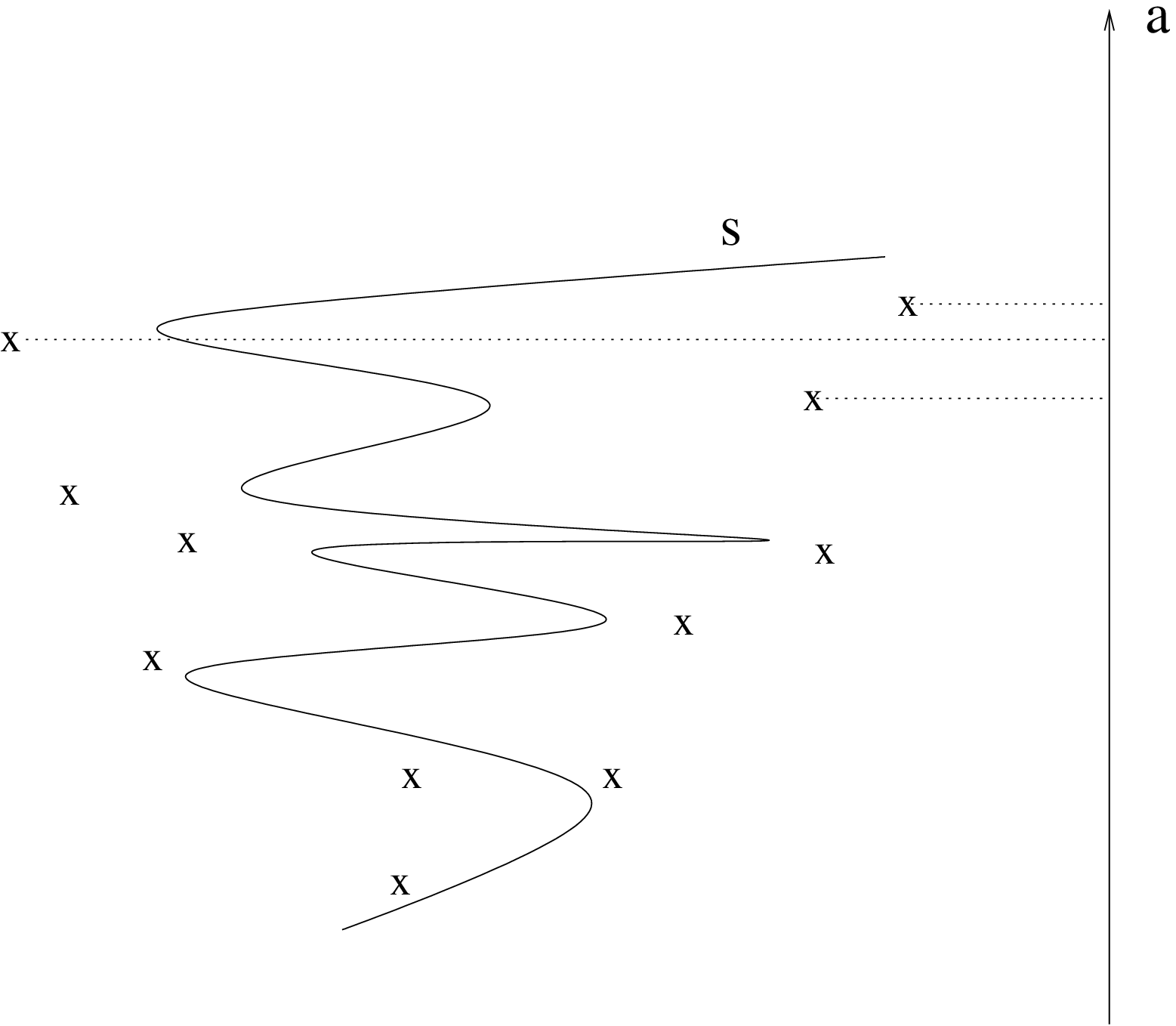}
\end{center}
\caption{Left: axis $a$ such that the associated projection $P_a$ preserves the first-order neighborhood structure.
The regression function $s$ correctly fits the dataset.
Right: axis $a$ for which $P_a$ does not preserve the first-order neighborhood structure.
The regression function $s$ cannot yield a good approximation of the dataset.}
\label{figaxe}     
\end{figure}

\noindent Our approach is similar to Lebart one's~\cite{lebart}.
It consists in defining a contiguity coefficient
whose minimization allows to unfold nonlinear structures.
At each iteration $k$, the following Rayleigh quotient~\cite{parlett}
is maximized with respect to~$x$:
\begin{equation}
\label{contigu}
I\circ P_{x}(R^{k-1}) =
{\sum_{i=1}^n P_{x}^2(R^{k-1}_i)}
\left/
{\sum_{i=1}^n\sum_{j=1}^n m_{i,j}^{k-1}P_{x}^2(R^{k-1}_i-R^{k-1}_j)}
\right..
\end{equation}
The matrix $M^{k-1}=(m_{i,j}^{k-1})$ is a first order contiguity matrix,
whose value is~$1$ when $R^{k-1}_j$ is the nearest neighbor of
$R^{k-1}_i$, $0$ otherwise.
The upper part of~(\ref{contigu}) is proportional to the usual projected
 variance, see~(\ref{indexpca}). 
The lower part is the distance between the projection of
points which are nearest neighbor in $\R^p$. Then, the maximization
of~(\ref{contigu}) should reveal directions in which the projection
best preserves the first order neighborhood structure (see Figure~\ref{figaxe}).
In this sense, the index~(\ref{contigu}) can be seen as a first order
approximation of the index proposed in~\cite{PAMI}.
Thanks to this approximation, the maximization step benefits
from an explicit solution:
The resulting principal direction $a^k$ 
is the eigenvector associated to the maximum
eigenvalue of $({V_{k-1}^\star})^{-1}V_{k-1}$ where
$$
V_{k-1}^\star= \frac{1}{n}\sum_{i=1}^n \sum_{j=1}^n m_{i,j}^{k-1}
{(R^{k-1}_i - R^{k-1}_j)\,^t(R^{k-1}_i - R^{k-1}_j)}
$$
is proportional to the local covariance matrix.
$({V_{k-1}^{\star}})^{-1}$ should be read as the generalized inverse
of the singular matrix $V_{k-1}^{\star}$. Indeed, since $R^{k-1}$
is orthogonal to
$\{a^1,\dots,a^{k-1}\}$ from~(\ref{eq1}), $V_{k-1}^\star$ is,
at most, of rank $p-k+1$.
Note that this approach is
equivalent to Lebart's one when the contiguity matrix $M$
is symmetric.

\section{Illustration on real and simulated data}
\label{secillus}

\begin{figure}
\centerline{\includegraphics[width=0.8\textwidth,angle=270]{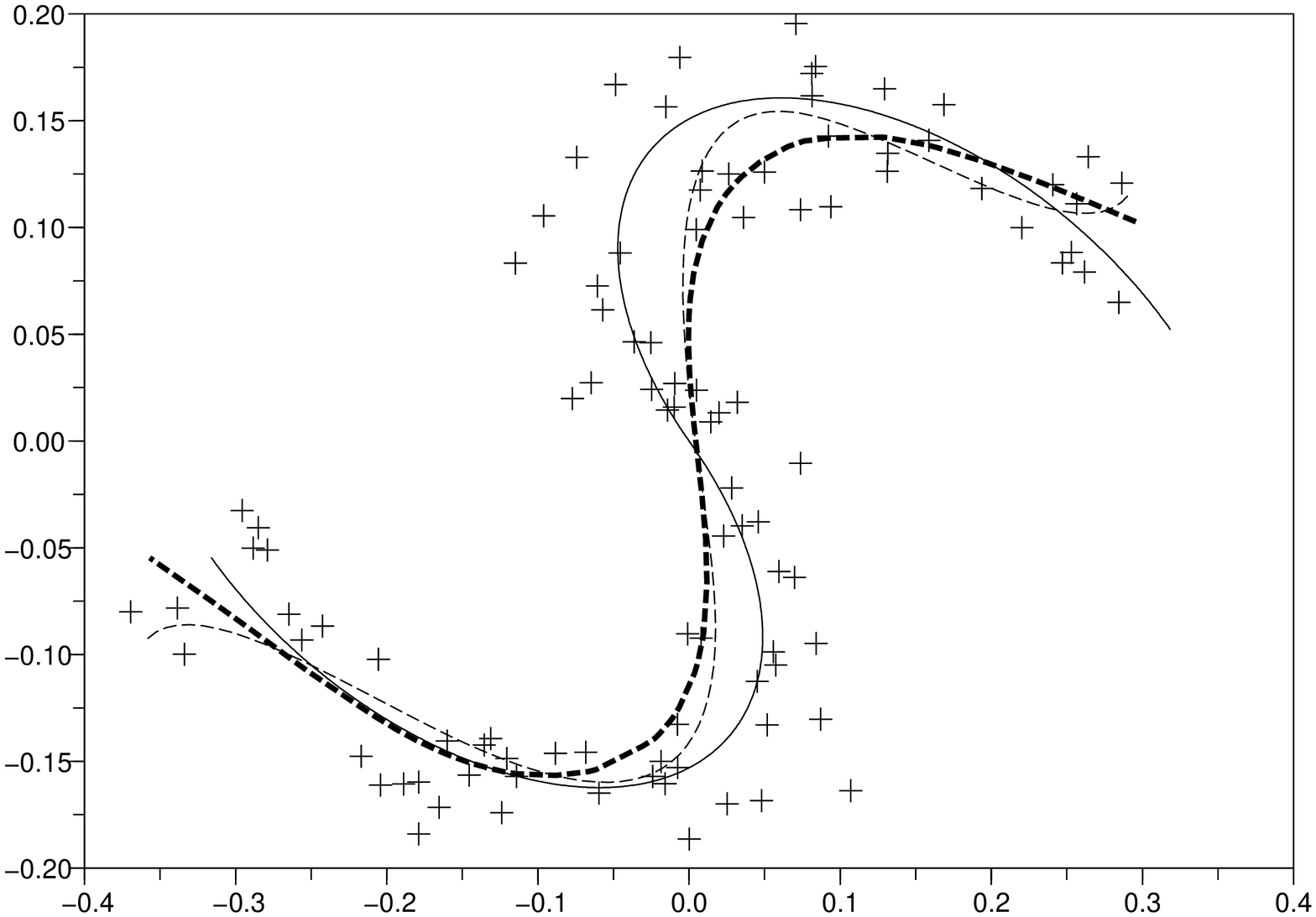}}
\caption{Comparison of one-dimensional estimated manifolds on a simulated dataset.
solid line: original curve, dashed line: curve estimated from the auto-associative model
approach, bold dashed line: principal curve estimated by the approach proposed in~\cite{HAS89}.}
\label{figkugl}
\centerline{\includegraphics[width=0.7\textwidth,angle=270]{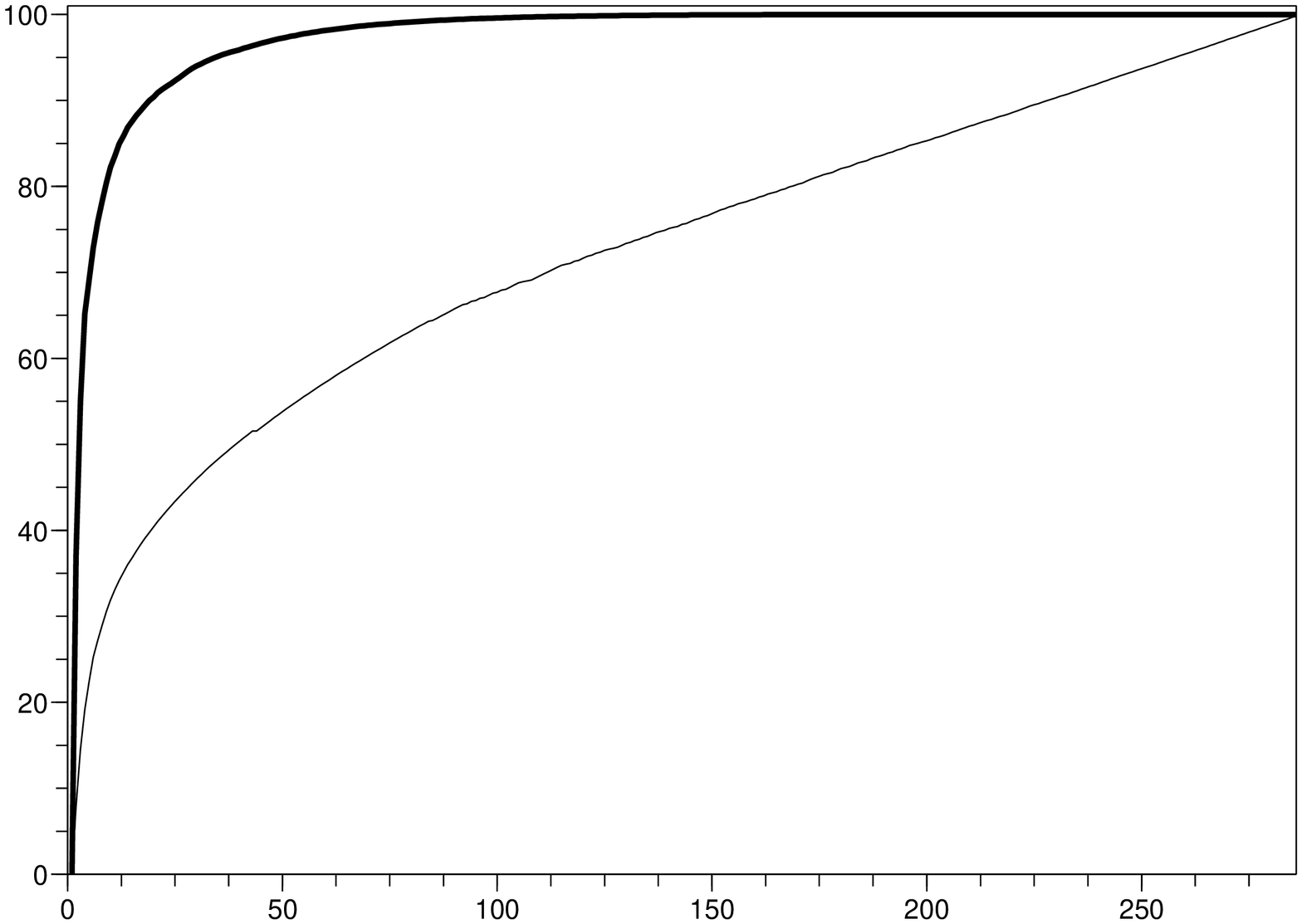}}
\caption{Forgetting the labels, information ratio $Q_d$ as a function of $d$ 
on a real dataset. solid line: classical PCA, bold line: generalized PCA based
on auto-associative models.}
\label{figqd}
\end{figure}

Our first illustration is done on the "DistortedSShape" simulated dataset
introduced in~\cite{theseKegl}, paragraph~5.2.1 and available on-line at the
following address:
{\tt http://www.iro.umontreal.ca/$\sim$kegl/research/pcurves}.\\
The dataset consists of 100 data points in $\R^2$ and located
around a one-dimensional curve (solid line on Figure~\ref{figkugl}).
The bold dashed curve is the one-dimensional manifold estimated by the
principal curves approach~\cite{HAS89}. The estimated curve
fails to follow the shape of the original curve.
Using the auto-associative model, the estimated one-dimensional
manifold (dashed curve) is closer to the original one.
In this experiment, we used one iteration of Algorithm~\ref{algoAA}
with the contiguity index~(\ref{contigu}) in combination
with a projection estimate of the regression functions.
A basis of $N=4$ cubic splines was used to compute the projection.

\begin{figure}
\centerline{\includegraphics[width=0.8\textwidth,angle=270]{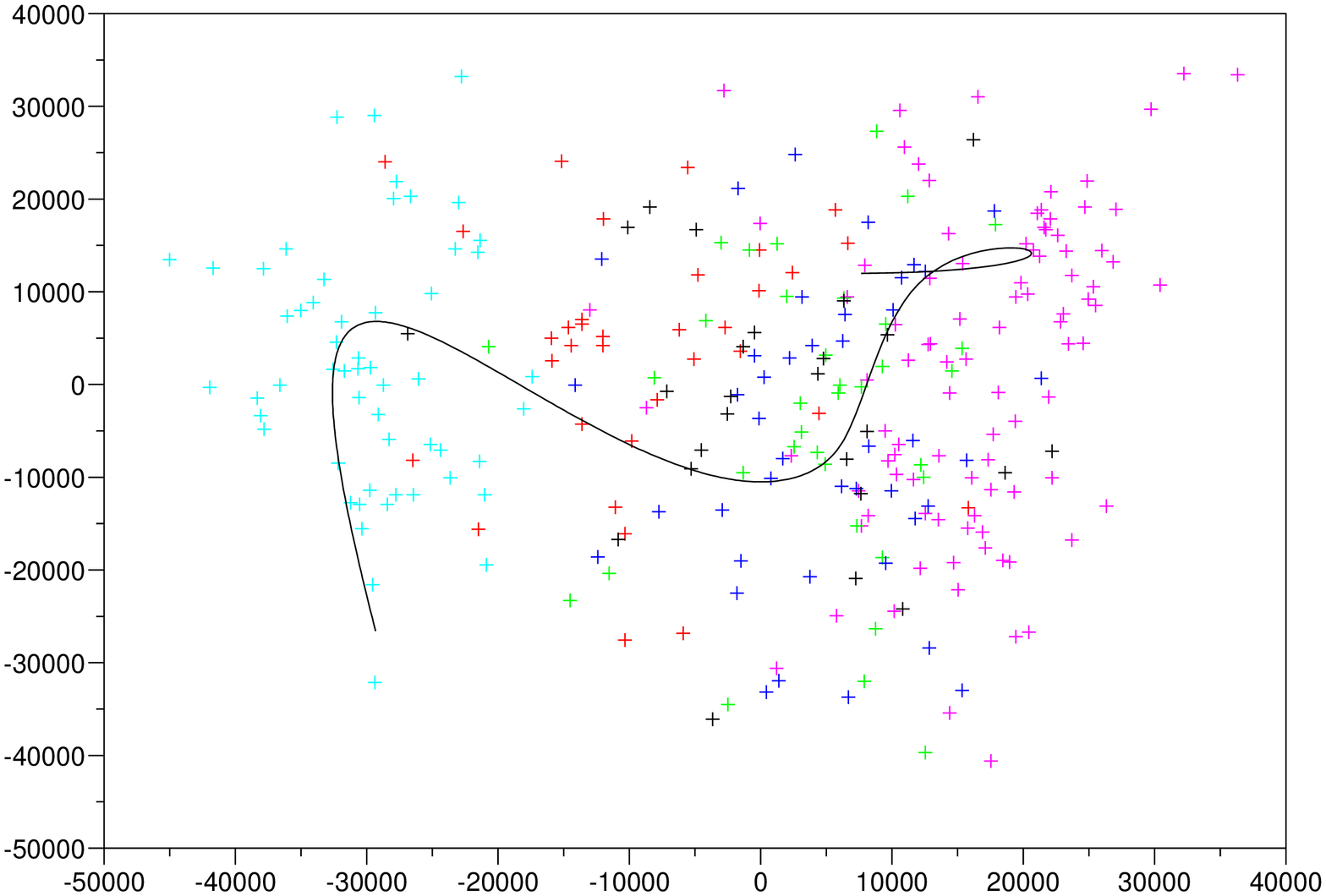}}
\caption{One-dimensional manifold estimated on a real dataset
with the auto-associative models approach and projected on the principal plane.}
\label{vardim1}
\centerline{\includegraphics[width=0.8\textwidth,angle=270]{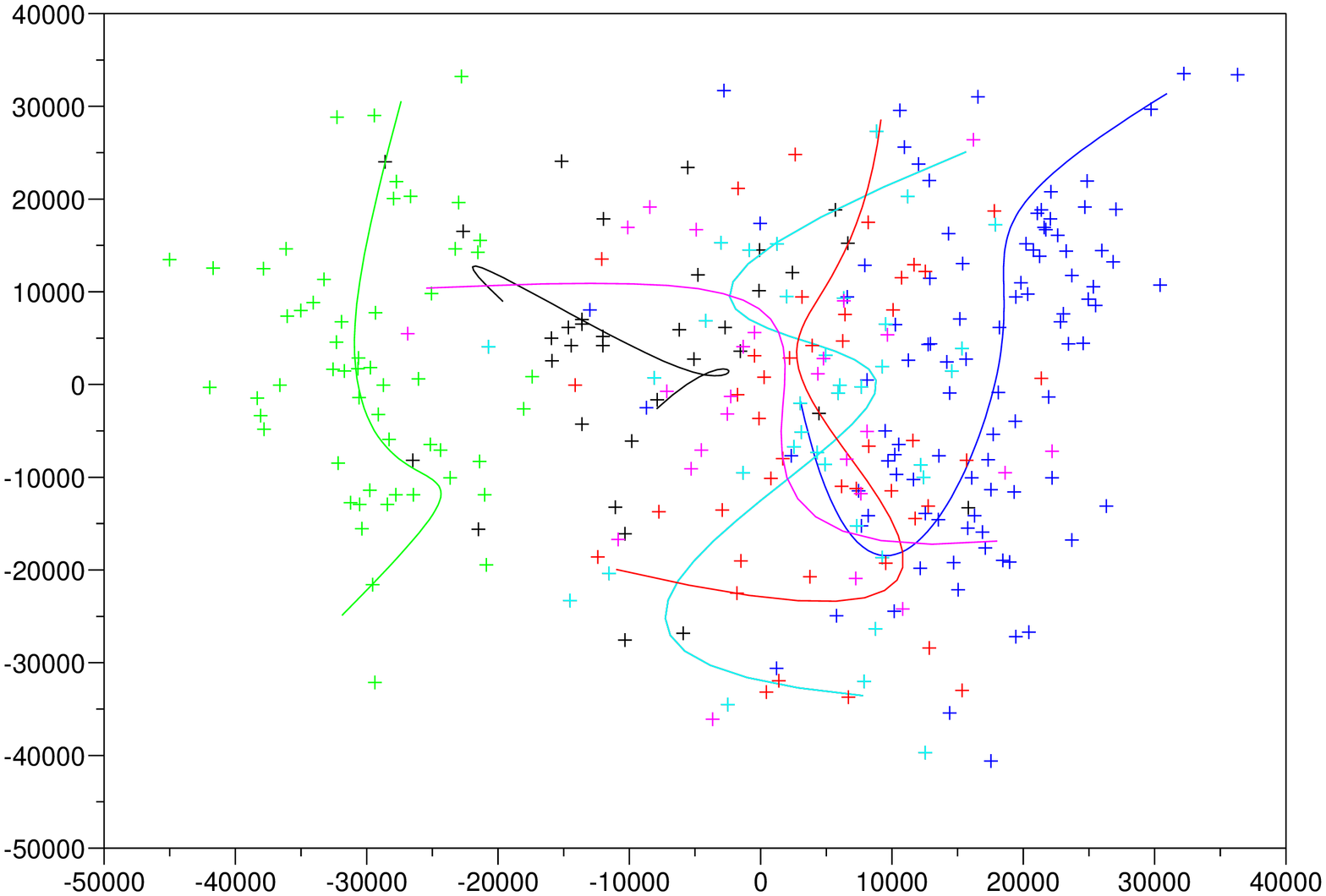}}
\caption{One-dimensional manifolds estimated on each type of cancer of the real dataset with the auto-associative models approach, and projected on the principal plane.}
\label{vardim1classes}
\end{figure}

\begin{figure}
\centerline{\includegraphics[width=1.2\textwidth]{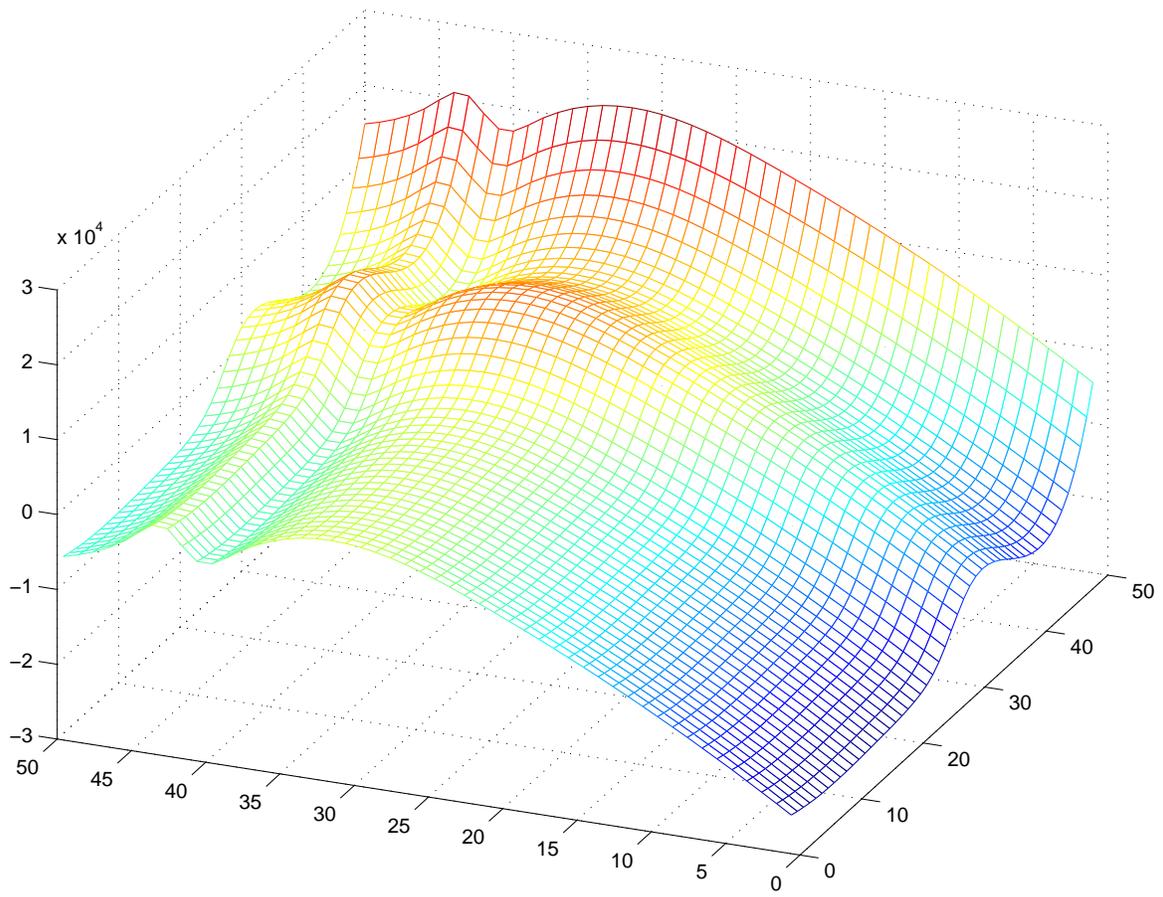}}
\caption{Two-dimensional manifold estimated on a real dataset
with the auto-associative models approach and projected on the three first
principal axes.}
\label{vardim2}
\end{figure}

Our second illustration is performed on the
"dataset I - Five types of breast cancer"
provided to us by the organizers of the 
"Principal Manifolds-2006" workshop.
The dataset is available on-line at the following address:
{\tt http://www.ihes.fr/$\sim$zinovyev/princmanif2006/}.\\
It consists of micro-array data
containing logarithms of expression levels of $p=17816$ genes in 
$n=286$ samples. 
The data is divided into five types of breast cancer 
(lumA, lumB, normal, errb2 and basal) plus an unclassified group.
Before all, let us note that, since $n$ points are
necessarily located on a linear subspace of dimension $n-1$,
the covariance matrix is at most of rank $n-1=285$.
Thus, as a preprocessing step, the dimension of the data
is reduced to $285$ by a classical PCA, and this, without
any loss of information.
Forgetting the labels, {\it i.e.} without using the initial
classification into five types of breast cancer,
the information ratio $Q_d$ (see Corollary~\ref{QD}) obtained
by the classical PCA and the generalized one (basing on auto-associative
models), are compared. Figure~\ref{figqd} illustrates the behavior of
$Q_d$ as the dimension $d$ of the model increases.
The bold curve, corresponding to the auto-associate model,
was computed with the contiguity index~(\ref{contigu}) in combination
with a projection estimate of the regression functions.
A basis of $N=2$ cubic splines was used to compute the projection.
One can see that the generalized PCA yields far better approximation
results than the classical one. 
As an illustration, the one-dimensional manifold is superimposed
to the dataset on Figure~\ref{vardim1}. Each class is represented
with a different gray level. For the sake of the visualization, the dataset
as well as the manifold are projected on the principal plane.
Similarly, the two-dimensional manifold is represented on Figure~\ref{vardim2}
on the linear space spanned by the three first principal axes.
Taking into account the labels, it is also possible to compute the
one-dimensional manifold associated to each type of cancer and
to the unclassified points,
see Figure~\ref{vardim1classes}. Each manifold then represents a kind
of skeleton of the corresponding dataset.

Other illustrations can be found in~\cite{Chalmond}, Chapter~4,
where auto-associative models are applied to some image analysis
problems.
 \bibliographystyle{plain}
 \bibliography{girard_iovleff3}

%
%
%

\end{document}